\documentclass{article}

\usepackage{arxiv}

\usepackage[utf8]{inputenc} % allow utf-8 input
\usepackage[T1]{fontenc}    % use 8-bit T1 fonts
\usepackage{hyperref}       % hyperlinks
\usepackage{url}            % simple URL typesetting
\usepackage{booktabs}       % professional-quality tables
\usepackage{amsfonts}       % blackboard math symbols
\usepackage{nicefrac}       % compact symbols for 1/2, etc.
\usepackage{microtype}      % microtypography
\usepackage{lipsum}
\usepackage{svg}
\usepackage{graphicx}
\graphicspath{ {./images/} }
\usepackage{amsmath} 
\usepackage{pdflscape}
\usepackage{xcolor}
\usepackage[normalem]{ulem}
\usepackage{comment}
 \usepackage{longtable}
\usepackage{tabularx}
\usepackage{caption} % for more complex captioning
\usepackage{subcaption}
\usepackage{longtable}
\usepackage{booktabs}
\usepackage{algorithm}
\usepackage{algpseudocode}
\usepackage{multirow}
\usepackage[utf8]{inputenc}

\title{A Review of YOLOv12: Attention-Based Enhancements vs. Previous Versions}

\author{
  \textbf{Rahima Khanam}\textsuperscript{*} and \textbf{Muhammad Hussain}\\[1ex] % Adds a little space between names and affiliation
  \begin{minipage}[t]{0.90\textwidth}
    \scriptsize Department of Computer Science, Huddersfield University, Queensgate, Huddersfield HD1 3DH, UK; \\
    \textsuperscript{*}Correspondence: rahima.khanam@hud.ac.uk;
  \end{minipage}
}
\begin{document}
\maketitle
\begin{abstract}
The YOLO (You Only Look Once) series has been a leading framework in real-time object detection, consistently improving the balance between speed and accuracy. However, integrating attention mechanisms into YOLO has been challenging due to their high computational overhead. YOLOv12 introduces a novel approach that successfully incorporates attention-based enhancements while preserving real-time performance. This paper provides a comprehensive review of YOLOv12’s architectural innovations, including Area Attention for computationally efficient self-attention, Residual Efficient Layer Aggregation Networks for improved feature aggregation, and FlashAttention for optimized memory access. Additionally, we benchmark YOLOv12 against prior YOLO versions and competing object detectors, analyzing its improvements in accuracy, inference speed, and computational efficiency. Through this analysis, we demonstrate how YOLOv12 advances real-time object detection by refining the latency-accuracy trade-off and optimizing computational resources. 
\end{abstract}


\section{Introduction}

Real-time object detection is a cornerstone of modern computer vision, playing a pivotal role in applications such as autonomous driving \cite{feng2020deep, feng2021review, mao20233d, lu2019review}, robotics \cite{manakitsa2024review, bai2020object, xu2022object}, and video surveillance \cite{joshi2018object, angadi2020review, mishra2016study}. These domains demand not only high accuracy but also low-latency performance to ensure real-time decision-making. Among the various object detection frameworks, the YOLO (You Only Look Once) series has emerged as a dominant solution \cite{redmon2016you}, striking a balance between speed and precision by continuously refining convolutional neural network (CNN) architectures \cite{redmon2017yolo9000, redmon2018yolov3, bochkovskiy2020yolov4, yolov5_ultralytics, li2022yolov6, wang2023yolov7, yolov8_ultralytics, wang2024yolov9, wang2024yolov10, yolo11_ultralytics}. However, a fundamental challenge in CNN-based detectors lies in their limited ability to capture long-range dependencies, which are crucial for understanding spatial relationships in complex scenes. This limitation has led to increased research into attention mechanisms, particularly Vision Transformers (ViTs) \cite{fang2024eva, he2022masked}, which excel at global feature modeling. Despite their advantages, ViTs suffer from quadratic computational complexity \cite{liu2025vmamba} and inefficient memory access \cite{dao2022flashattention, dao2023flashattention}, making them impractical for real-time deployment. 

To address these limitations, YOLOv12 \cite{tian2025yolov12} introduces an attention-centric approach that integrates key innovations to enhance efficiency while maintaining real-time performance. By embedding attention mechanisms within the YOLO framework, it successfully bridges the gap between CNN-based and transformer-based detectors without compromising speed. This is achieved through several architectural enhancements that optimize computational efficiency, improve feature aggregation, and refine attention mechanisms:  

\begin{enumerate}
    \item \textbf{Area Attention (A2)}: A novel mechanism that partitions spatial regions to reduce the complexity of self-attention, preserving a large receptive field while improving computational efficiency. This enables attention-based models to compete with CNNs in speed.
    \item \textbf{Residual Efficient Layer Aggregation Networks (R-ELAN)}: An enhancement over traditional ELAN, designed to stabilize training in large-scale models by introducing residual shortcuts and a revised feature aggregation strategy, ensuring better gradient flow and optimization.
    \item \textbf{Architectural Streamlining}: Several structural refinements, including the integration of FlashAttention for efficient memory access, the removal of positional encoding to simplify computations, and an optimized MLP ratio to balance performance and inference speed.
\end{enumerate}

This review systematically examines the key architectural advancements in YOLOv12, including the integration of attention mechanisms, feature aggregation strategies, and computational optimizations. To provide a structured analysis, the paper is organized as follows: Section \ref{sec:TE} outlines the technical evolution of YOLO architectures, highlighting the advancements leading to YOLOv12. Section \ref{sec:AD} details the architectural design of YOLOv12, describing its backbone, feature extraction process, and detection head. Section \ref{sec:AI} explores the model’s key innovations, including the A2 module, R-ELAN, and additional enhancements for improved efficiency. Section \ref{sec:BE} presents a benchmark evaluation, comparing YOLOv12’s performance with previous YOLO versions and state-of-the-art object detectors. Section \ref{sec:CVT} discusses the various computer vision tasks supported by YOLOv12. Section \ref{sec:D} provides a broader discussion on model efficiency, deployment considerations, and the impact of YOLOv12 in real-world applications. Section \ref{sec:CFR} addresses current challenges and outlines future research directions. Finally, Section \ref{sec:C} concludes the paper by summarizing YOLOv12’s contributions to real-time object detection and its potential for further advancements in the field.

%%%%%%%%%%%%%%%%%%%%%%%%%%%%%%%%%%%%%%%%%

\section{Technical Evolution of YOLO Architectures} \label{sec:TE}

The \textit{You Only Look Once }(YOLO) series has revolutionized real-time object detection through continuous architectural innovation and performance optimization. The evolution of YOLO can be traced through distinct versions, each introducing significant advancements.

\textbf{YOLOv1 (2015)} \cite{redmon2016you}, developed by Joseph Redmon et al., introduced the concept of single-stage object detection, prioritizing speed over accuracy. It divided the image into a grid and predicted bounding boxes and class probabilities directly from each grid cell, enabling real-time inference. This method significantly reduced the computational overhead compared to two-stage detectors, albeit with some trade-offs in localization accuracy.

\textbf{YOLOv2 (2016)} \cite{redmon2017yolo9000}, also by Joseph Redmon, enhanced detection capabilities with the introduction of anchor boxes, batch normalization, and multi-scale training. Anchor boxes allowed the model to predict bounding boxes of various shapes and sizes, improving its ability to detect diverse objects. Batch normalization stabilized training and improved convergence, while multi-scale training made the model more robust to varying input resolutions.

\textbf{YOLOv3 (2018)} \cite{redmon2018yolov3}, again by Joseph Redmon, further improved accuracy with the Darknet-53 backbone, Feature Pyramid Networks (FPN), and logistic classifiers. Darknet-53 provided a deeper and more powerful feature extractor, while FPN enabled the model to leverage multi-scale features for improved detection of small objects. Logistic classifiers replaced softmax for class prediction, allowing for multi-label classification.

\textbf{YOLOv4 (2020)} \cite{bochkovskiy2020yolov4}, developed by Alexey Bochkovskiy et al., incorporated CSPDarknet, Mish activation, PANet, and Mosaic augmentation. CSPDarknet reduced computational costs while maintaining performance, Mish activation improved gradient flow, PANet enhanced feature fusion, and Mosaic augmentation increased data diversity.

\textbf{YOLOv5 (2020)} \cite{yolov5_ultralytics}, developed by Ultralytics, marked a pivotal shift by introducing a PyTorch implementation. This significantly simplified training and deployment, making YOLO more accessible to a wider audience. It also featured auto-anchor learning, which dynamically adjusted anchor box sizes during training, and incorporated advancements in data augmentation. The transition from Darknet to PyTorch was a major change, and greatly contributed to the models popularity.

\textbf{YOLOv6 (2022)} \cite{li2022yolov6}, developed by Meituan, focused on efficiency with the EfficientRep backbone, Neural Architecture Search (NAS), and RepOptimizer. EfficientRep optimized the model's architecture for speed and accuracy, NAS automated the search for optimal hyperparameters, and RepOptimizer reduced inference time through structural re-parameterization.

\textbf{YOLOv7 (2022) }\cite{wang2023yolov7}, developed by Wang et al., further improved efficiency through Extended Efficient Layer Aggregation Network (E-ELAN) and re-parameterized convolutions. E-ELAN enhanced feature integration and learning capacity, while re-parameterized convolutions reduced computational overhead.

\textbf{YOLOv8 (2023)} \cite{yolov8_ultralytics}, also developed by Ultralytics, introduced C2f modules, task-specific detection heads, and anchor-free detection. C2f modules enhanced feature fusion and gradient flow, task-specific detection heads allowed for more specialized detection tasks, and anchor-free detection eliminated the need for predefined anchor boxes, simplifying the detection process.

\textbf{YOLOv9 (2024)} \cite{wang2024yolov9}, developed by Chien-Yao Wang et al., introduces Generalized Efficient Layer Aggregation Network (GELAN) and Programmable Gradient Information (PGI). GELAN improves the models ability to learn diverse features, and PGI helps to avoid information loss during deep network training.

\textbf{YOLOv10 (2024)} \cite{wang2024yolov10}, developed by various research contributors, emphasizes dual label assignments, NMS-free detection, and end-to-end training. Dual label assignments enhance the model's ability to handle ambiguous object instances, NMS-free detection reduces computational overhead, and end-to-end training simplifies the training process. The reason for stating "various research contributors" is that, at this time, there isn't a single, universally recognized, and consistently credited developer or organization for this specific release, as with previous versions.

\textbf{YOLOv11 (2024)} \cite{yolo11_ultralytics}, developed by Glenn Jocher and Jing Qiu, focuses on the C3K2 module, feature aggregation, and optimized training pipelines. The C3K2 module enhances feature extraction, feature aggregation improves the model's ability to integrate multi-scale features, and optimized training pipelines reduce training time. Similar to YOLOv10, the developer information is less consolidated and more collaborative.

\textbf{YOLOv12 (2025)} \cite{tian2025yolov12}, the latest iteration, integrates attention mechanisms while preserving real-time efficiency. It introduces A2, Residual-Efficient Layer Aggregation Networks (R-ELAN), and FlashAttention, alongside a hybrid CNN-Transformer framework. These innovations refine computational efficiency and optimize the latency-accuracy trade-off, surpassing both CNN-based and transformer-based object detectors.

\begin{figure}[h]
    \centering
    \includegraphics[width=\linewidth]{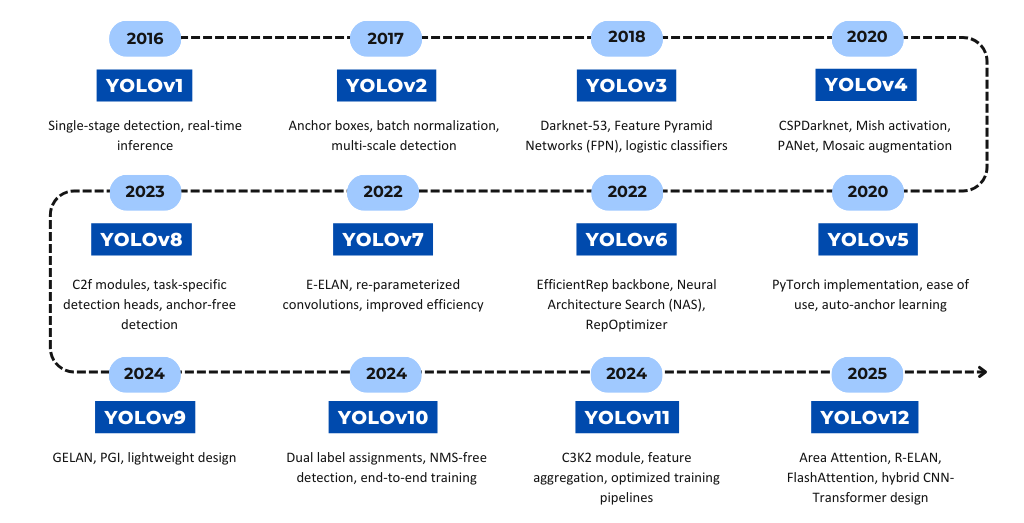}
    \caption{Evolution of YOLO architectures}
    \label{fig:yolo-evol}
\end{figure}

The evolution of YOLO models highlights a shift from Darknet-based architectures \cite{redmon2016you, redmon2017yolo9000, redmon2018yolov3, bochkovskiy2020yolov4} to PyTorch implementations \cite{yolov5_ultralytics, li2022yolov6, wang2023yolov7, yolov8_ultralytics, wang2024yolov9, wang2024yolov10, yolo11_ultralytics}, and more recently, towards hybrid CNN-transformer architectures \cite{tian2025yolov12}. Each generation has balanced speed and accuracy, incorporating advancements in feature extraction, gradient optimization, and data efficiency. Figure \ref{fig:yolo-evol} illustrates the progression of YOLO architectures, emphasizing key innovations across versions.

With YOLOv12’s architectural refinements, attention mechanisms are now embedded within the YOLO framework, optimizing both computational efficiency and high-speed inference. The next section analyzes these enhancements in detail, benchmarking YOLOv12’s performance across multiple detection tasks.

%%%%%%%%%%%%%%%%%%%%%%%%%%%%%%%%%%%%%%%%%%

\section{Architectural Design of YOLOv12} \label{sec:AD}

The YOLO framework revolutionized object detection by introducing a unified neural network that simultaneously performs bounding box regression and object classification in a single forward pass \cite{khanam2024comprehensive}. Unlike traditional two-stage detection methods, YOLO adopts an end-to-end approach, making it highly efficient for real-time applications. Its fully differentiable design allows seamless optimization, leading to improved speed and accuracy in object detection tasks.

At its core, the YOLOv12 architecture consists of two primary components: the backbone and the head. The backbone serves as the feature extractor, processing the input image through a series of convolutional layers to generate hierarchical feature maps at different scales. These features capture essential spatial and contextual information necessary for object detection. The head is responsible for refining these features and generating final predictions by performing multi-scale feature fusion and localization. Through a combination of upsampling, concatenation, and convolutional operations, the head enhances feature representations, ensuring robust detection of small, medium, and large objects. The Backbone and Head Architecture of YOLOv12 is depicted in Algorithm 1.

\begin{algorithm}
\caption{Backbone and Head Architecture of YOLOv12}
\begin{algorithmic}[]

\State \textbf{Input:} Image $I$
\State \textbf{Output:} Detection predictions

\Procedure{Backbone }{I}
    \State \textbf{Parameters:} $nc = 80$ \Comment{Number of classes}
    \State \textbf{Scales:} $[0.50, 0.25, 1024], [0.50, 0.50, 1024], [0.50, 1.00, 512], [1.00, 1.00, 512], [1.00, 1.50, 512]$
    
    \State \textbf{/* Feature Extraction */}
    \State $P1 \gets \text{Conv}(I, 64, 3, 2)$  \Comment{P1/2}
    \State $P2 \gets \text{Conv}(P1, 128, 3, 2)$  \Comment{P2/4}
    \State $P2 \gets \text{C3k2}(P2, 256, \text{False}, 0.25)$
    \State $P3 \gets \text{Conv}(P2, 256, 3, 2)$  \Comment{P3/8}
    \State $P3 \gets \text{C3k2}(P3, 512, \text{False}, 0.25)$
    \State $P4 \gets \text{Conv}(P3, 512, 3, 2)$  \Comment{P4/16}
    \State $P4 \gets \text{A2C2F}(P4, 512, \text{True}, 4)$
    \State $P5 \gets \text{Conv}(P4, 1024, 3, 2)$  \Comment{P5/32}
    \State $P5 \gets \text{A2C2F}(P5, 1024, \text{True}, 1)$
    
    \Return $P3, P4, P5$
\EndProcedure

\Procedure{Head }{P3, P4, P5}
    \State \textbf{/* Feature Fusion and Upsampling */}
    \State $U1 \gets \text{Upsample}(P5, \text{"nearest"})$
    \State $C1 \gets \text{Concat}([U1, P4])$ \Comment{Merge P5 with P4}
    \State $H1 \gets \text{A2C2F}(C1, 512, \text{False})$

    \State $U2 \gets \text{Upsample}(H1, \text{"nearest"})$
    \State $C2 \gets \text{Concat}([U2, P3])$ \Comment{Merge P4 with P3}
    \State $H2 \gets \text{A2C2F}(C2, 256, \text{False})$

    \State \textbf{/* Detection Head Processing */}
    \State $H3 \gets \text{Conv}(H2, 256, 3, 2)$
    \State $C3 \gets \text{Concat}([H3, P4])$ \Comment{Merge P3 with P4}
    \State $H4 \gets \text{A2C2F}(C3, 512, \text{False})$

    \State $H5 \gets \text{Conv}(H4, 512, 3, 2)$
    \State $C4 \gets \text{Concat}([H5, P5])$ \Comment{Merge P4 with P5}
    \State $H6 \gets \text{C3k2}(C4, 1024, \text{True})$ \Comment{P5/32-large}

    \State \textbf{/* Final Detection */}
    \State $D \gets \text{Detect}([H2, H4, H6], nc)$

    \Return $D$
\EndProcedure

\end{algorithmic}
\end{algorithm}

\subsection{Backbone: Feature Extraction}
The backbone of YOLOv12 processes the input image through a series of convolutional layers, progressively reducing its spatial dimensions while increasing the depth of feature maps. The process begins with an initial convolutional layer that extracts low-level features, followed by additional convolutional layers that perform downsampling to capture hierarchical information. The first stage applies a 3×3 convolution with a stride of 2 to generate the initial feature map. This is followed by another convolutional layer that further reduces the spatial resolution while increasing feature depth.

As the image moves through the backbone, it undergoes multi-scale feature learning using specialized modules like C3k2 and A2C2F. The C3k2 module enhances feature representation while maintaining computational efficiency, and the A2C2F module improves feature fusion for better spatial and contextual understanding. The backbone continues this process until it generates three key feature maps: P3, P4, and P5, each representing different scales of feature extraction. These feature maps are then passed to the detection head for further processing.

\subsection{Head: Feature Fusion and Object Detection}
The head of YOLOv12 is responsible for merging multi-scale features and generating final object detection predictions. It employs a feature fusion strategy that combines information from different levels of the backbone to enhance detection accuracy across small, medium, and large objects. This is achieved through a series of upsampling and concatenation operations. The process begins with the highest-resolution feature map (P5) being upsampled using a nearest-neighbor interpolation method. It is then concatenated with the corresponding lower-resolution feature map (P4) to create a refined feature representation. The fused feature is further processed using the A2C2F module to enhance its expressiveness.

A similar process is repeated for the next scale by upsampling the refined feature map and concatenating it with the lower-scale feature (P3). This hierarchical fusion ensures that both low-level and high-level features contribute to the final detection, improving the model’s ability to detect objects at varying scales.

After feature fusion, the network undergoes final processing to prepare for detection. The refined features are downsampled again and merged at different levels to strengthen object representations. The C3k2 module is applied at the largest scale (P5/32-large) to ensure that high-resolution features are preserved while reducing computational cost. These processed feature maps are then passed through the final detection layer, which applies classification and localization predictions across different object categories. The detailed breakdown of its backbone and head architecture is formally described in Algorithm 1.

\section{Architectural Innovations of YOLOv12} \label{sec:AI}

YOLOv12 introduces a novel attention-centric approach to real-time object detection, bridging the performance gap between conventional CNNs and attention-based architectures. Unlike previous YOLO versions that primarily relied on CNNs for efficiency, YOLOv12 integrates attention mechanisms without sacrificing speed. This is achieved through three key architectural improvements: the A2 Module, R-ELAN, and enhancements to the overall model structure, including FlashAttention and reduced computational overhead in the multi-layer perceptron (MLP). Each of these components is detailed below:

\subsection{Area Attention Module}

The efficiency of attention mechanisms has traditionally been hindered by their high computational cost, particularly due to the quadratic complexity associated with self-attention operations \cite{wang2020linformer}. A common strategy to mitigate this issue is linear attention \cite{shen2021efficient}, which reduces complexity by approximating attention interactions with more efficient transformations. However, while linear attention improves speed, it suffers from global dependency degradation \cite{katharopoulos2020transformers}, instability during training \cite{choromanski2020rethinking}, and sensitivity to input distribution shifts \cite{xiong2021nystromformer}. Additionally, due to its low-rank representation constraints \cite{bhojanapalli2020low, choromanski2020rethinking}, it struggles to retain fine-grained details in high-resolution images, limiting its effectiveness in object detection. 

\begin{figure}[h]
    \centering
    \includegraphics[width=0.7\textwidth]{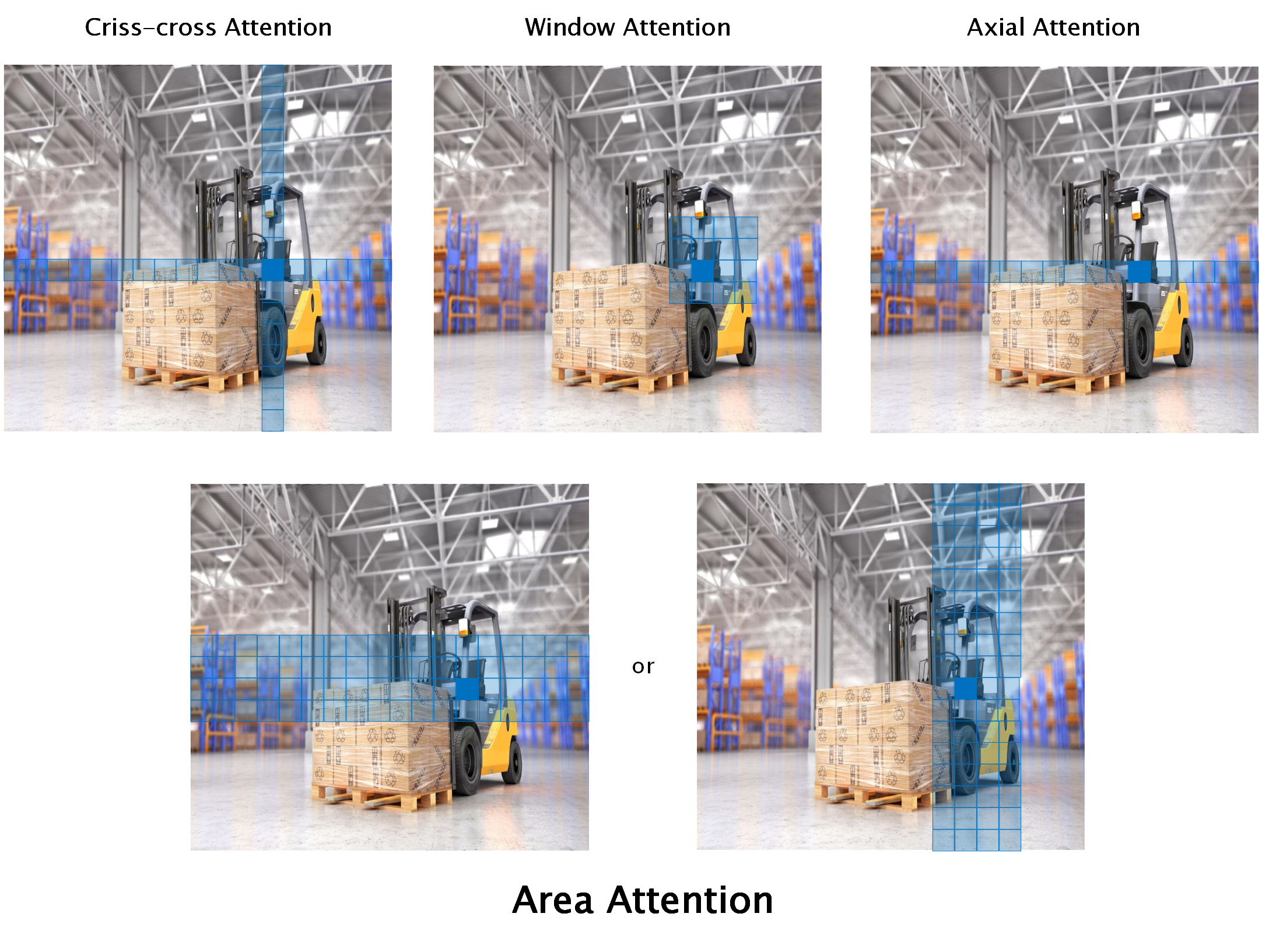}
    \caption{Comparison of different local attention techniques, with the proposed Area Attention method}
    \label{fig:locatten_comp}
\end{figure}

To address these limitations, YOLOv12 introduces the A2 Module, which retains the strengths of self-attention while significantly reducing computational overhead \cite{tian2025yolov12}. Unlike traditional global attention mechanisms that compute interactions across the entire image, Area Attention divides the feature map into equal-sized non-overlapping segments, either horizontally or vertically. Specifically, a feature map of dimensions \((H, W)\) is partitioned into \(L\) segments of size \((H/L, W)\) or \((H, W/L)\), eliminating the need for explicit window partitioning methods seen in other attention models such as Shifted Window \cite{liu2021swin}, Criss-Cross Attention \cite{huang2019ccnet}, or Axial Attention \cite{dong2022cswin}. These methods often introduce additional complexity and reduce computational efficiency, whereas A2 achieves segmentation via a simple reshape operation, maintaining a large receptive field while significantly enhancing processing speed \cite{tian2025yolov12}. This approach is depicted in Figure \ref{fig:locatten_comp}.

Although A2 reduces the receptive field to \( \frac{1}{4} \) of the original size, it still surpasses conventional local attention methods in coverage and efficiency. Moreover, its computational cost is nearly halved, reducing from \( 2n^2hd \) (traditional self-attention complexity) to \( \frac{n^2hd}{2} \). This efficiency gain allows YOLOv12 to process large-scale images more effectively while maintaining robust detection accuracy \cite{tian2025yolov12}.  

%%%%%

\subsection{Residual Efficient Layer Aggregation Networks (R-ELAN)}

Feature aggregation plays a crucial role in improving information flow within deep learning architectures. Previous YOLO models incorporated Efficient Layer Aggregation Networks (ELAN) \cite{wang2023yolov7}, which optimized feature fusion by splitting the output of $1 \times 1$ convolution layers into multiple parallel processing streams before merging them back together. However, this approach introduced two major drawbacks: \textit{gradient blocking} and \textit{optimization difficulties}. These issues were particularly evident in deeper models, where the lack of direct residual connections between the input and output impeded effective gradient propagation, leading to slow or unstable convergence.

To address these challenges, YOLOv12 introduces R-ELAN, a novel enhancement designed to improve training stability and convergence. Unlike ELAN, R-ELAN integrates \textit{residual shortcuts} that connect the input directly to the output with a scaling factor (default set to 0.01) \cite{tian2025yolov12}. This ensures smoother gradient flow while maintaining computational efficiency. These residual connections are inspired by layer scaling techniques in Vision Transformers \cite{touvron2021going}, but they are specifically adapted to convolutional architectures to prevent latency overhead, which often affects attention-heavy models.

\begin{figure}[h]
    \centering
    \includegraphics[width=0.9\textwidth]{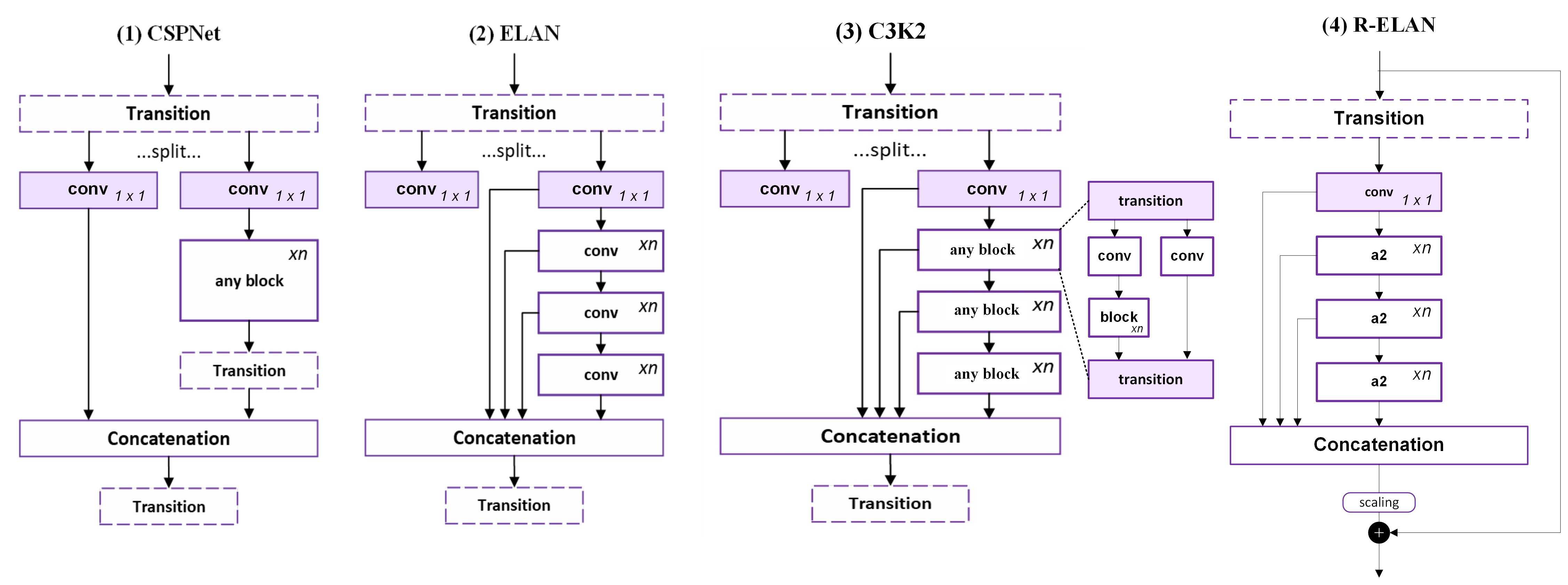}
    \caption{Comparison of CSPNet, ELAN, C3k2, and R-ELAN Architectures.}
    \label{fig:arch_comparison}
\end{figure}

Figure~\ref{fig:arch_comparison} illustrates a comparative overview of different architectures, including CSPNet, ELAN, C3k2, and R-ELAN, highlighting their structural distinctions.

\begin{itemize}
    \item \textbf{CSPNet (Cross-Stage Partial Network)}: CSPNet improves gradient flow and reduces redundant computation by splitting the feature map into two parts, processing one through a sequence of convolutions while keeping the other unaltered, and then merging them. This partial connection approach enhances efficiency while preserving representational capacity \cite{wang2020cspnet}.
    
    \item \textbf{ELAN (Efficient Layer Aggregation Networks)}: ELAN extends CSPNet by introducing deeper feature aggregation. It utilizes multiple parallel convolutional paths after the initial $1 \times 1$ convolution, which are concatenated to enrich feature representation. However, the absence of direct residual connections limits gradient flow, making deeper networks harder to train \cite{wang2023yolov7}.
    
    \item \textbf{C3k2}: A modified version of ELAN, C3k2 incorporates additional transformations within the feature aggregation process, but it still inherits the gradient-blocking issues from ELAN. While it improves structural efficiency, it does not fully resolve the optimization challenges faced in deep networks \cite{yolo11_ultralytics, wang2024yolov9}.
    
    \item \textbf{R-ELAN}: Unlike ELAN and C3k2, R-ELAN restructures feature aggregation by incorporating residual connections. Instead of first splitting the feature map and processing the parts independently, R-ELAN adjusts channel dimensions upfront, generating a unified feature map before passing it through bottleneck layers. This design significantly enhances computational efficiency by reducing redundant operations while ensuring effective feature integration \cite{tian2025yolov12}.
\end{itemize}

The introduction of R-ELAN in YOLOv12 yields several advantages, including faster convergence, improved gradient stability, and reduced optimization difficulties, particularly for larger-scale models (L- and X-scale). Previous versions often faced convergence failures under standard optimizers like Adam and AdamW  \cite{wang2023yolov7}, but R-ELAN effectively mitigates these issues, making YOLOv12 more robust for deep learning applications \cite{tian2025yolov12}.

\subsection{Additional Improvements and Efficiency Enhancements}

Beyond the introduction of A2 and R-ELAN, YOLOv12 incorporates several additional architectural refinements to enhance overall performance:

\begin{itemize}
    \item \textbf{Streamlined Backbone with Fewer Stacked Blocks:} Prior versions of YOLO \cite{yolov8_ultralytics, wang2024yolov9, wang2024yolov10, yolo11_ultralytics} incorporated multiple stacked attention and convolutional layers in the final stages of the backbone. YOLOv12 optimizes this by retaining only a single R-ELAN block, leading to faster convergence, better optimization stability, and improved inference efficiency—especially in larger models.
    
    \item \textbf{Efficient Convolutional Design:} To enhance computational efficiency, YOLOv12 strategically retains convolution layers where they offer advantages. Instead of using fully connected layers with Layer Normalization (LN), it adopts convolution operations combined with Batch Normalization (BN), which better suits real-time applications \cite{tian2025yolov12}. This allows the model to maintain CNN-like efficiency while incorporating attention mechanisms.

    \item \textbf{Removal of Positional Encoding:} Unlike traditional attention-based architectures, YOLOv12 discards explicit positional encoding and instead employs large-kernel separable convolutions (7×7) in the attention module \cite{tian2025yolov12}, known as the Position Perceiver. This ensures spatial awareness without adding unnecessary complexity, improving both efficiency and inference speed.

    \item \textbf{Optimized MLP Ratio:} Traditional Vision Transformers typically use an MLP expansion ratio of 4, leading to computational inefficiencies when deployed in real-time settings. YOLOv12 reduces the MLP ratio to 1.2 \cite{tian2025yolov12}, ensuring that the feed-forward network does not dominate overall runtime. This refinement helps balance efficiency and performance, preventing unnecessary computational overhead.
    
    \item \textbf{FlashAttention Integration:} One of the key bottlenecks in attention-based models is memory inefficiency \cite{dao2022flashattention, dao2023flashattention}. YOLOv12 incorporates FlashAttention, an optimization technique that reduces memory access overhead by restructuring computation to better utilize GPU high-speed memory (SRAM). This allows YOLOv12 to match CNNs in terms of speed while leveraging the superior modeling capacity of attention mechanisms.

\end{itemize}

\section{Benchmark Evaluation of YOLOv12} \label{sec:BE}

\begin{figure}[h]
    \centering
    \includegraphics[width=1\linewidth]{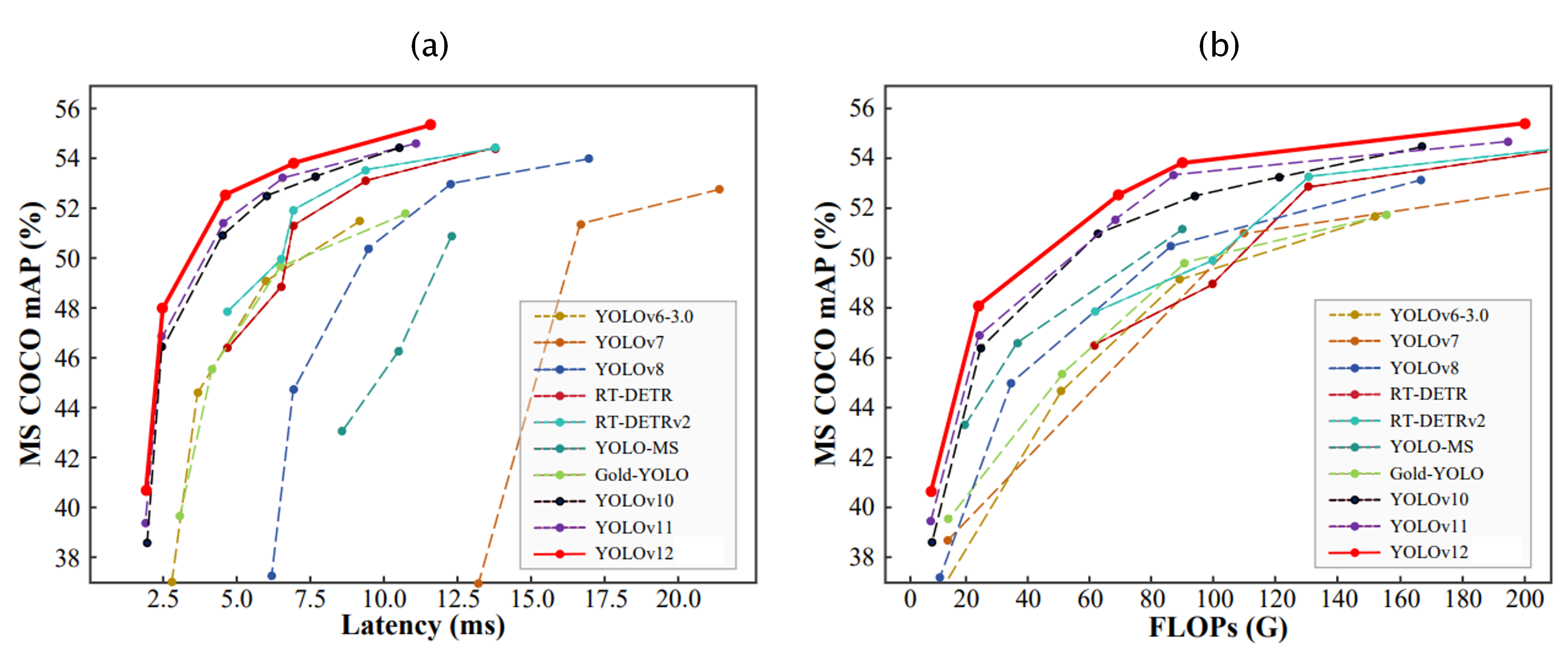}
    \caption{Benchmark comparison of YOLOv12 against prior models. (a) mAP vs. Latency. (b) mAP vs. FLOPs \cite{tian2025yolov12}.}
    \label{fig:yolov12-vs-prev}
\end{figure}

Evaluating the performance of object detection models requires a comprehensive analysis of both accuracy and computational efficiency. YOLOv12 is assessed on the MS COCO 2017 object detection benchmark \cite{lin2014microsoft}, a standard dataset used to evaluate object detection models. Its performance is compared against previous YOLO versions and state-of-the-art detection models, including RT-DETR and RT-DETRv2. The evaluation considers key metrics such as mean Average Precision (mAP), inference latency, and FLOPs, providing insights into YOLOv12’s effectiveness in real-world applications. The results are visualized in Figure \ref{fig:yolov12-vs-prev} and are detailed in the following sections, highlighting YOLOv12’s advancements in accuracy, speed, and computational efficiency. 

\subsection{Latency vs. Accuracy}  
Inference speed is a critical factor in real-time object detection applications, where responsiveness is paramount. The results in Figure~\ref{fig:yolov12-vs-prev} (a) demonstrate that YOLOv12 achieves higher mAP than previous YOLO models while maintaining competitive or superior latency. For instance, the smallest variant, YOLOv12-N, attains 40.6\% mAP, surpassing YOLOv10-N (38.5\%) and YOLOv11-N (39.4\%), with a comparable inference time of 1.64 ms on a T4 GPU. The larger YOLOv12-X model achieves 55.2\% mAP, outperforming its predecessor YOLOv11-X by 0.6\%, demonstrating the effectiveness of the model refinements in both accuracy and computational efficiency. This consistent improvement across model sizes underscores the efficacy of YOLOv12's architecture and optimization strategies.

Notably, YOLOv12 maintains a consistent advantage over RT-DETR models, particularly in inference speed. YOLOv12-S runs approximately 42\% faster than RT-DETR-R18/RT-DETRv2-R18, while utilizing only 36\% of the computation and 45\% of the parameters. Specifically, YOLOv12-S achieves a latency of 2.61 ms compared to 4.58 ms for RT-DETR-R18/RT-DETRv2-R18, highlighting a significant speed advantage. These improvements highlight the efficiency of YOLOv12 in reducing latency while preserving or enhancing detection accuracy, making it exceptionally well-suited for time-sensitive applications such as autonomous driving, surveillance, and robotics, where rapid processing is crucial. 

\subsection{FLOPs vs. Accuracy}  
Figure~\ref{fig:yolov12-vs-prev} (b) illustrates the relationship between mAP and FLOPs (floating-point operations per second), providing detailed insights into the computational efficiency of YOLOv12. The results indicate that YOLOv12 achieves higher accuracy at comparable or lower FLOPs than competing architectures. The red curve, representing YOLOv12, consistently remains above competing models, demonstrating that YOLOv12 effectively utilizes computational resources to maximize accuracy. This efficient utilization is pivotal for deploying models on devices with limited computational power.

A key observation is that YOLOv12 scales efficiently across different model sizes. While increasing FLOPs typically leads to higher accuracy, YOLOv12 consistently outperforms prior models with the same or fewer FLOPs, reinforcing the benefits of its architectural optimizations. For example, YOLOv12-L achieves 53.7\% mAP with 88.9 GFLOPs, surpassing YOLOv11-L which achieves 53.3\% mAP with 86.9 GFLOPs. This trend suggests that YOLOv12 can maintain high efficiency even under computational constraints, making it suitable for deployment on resource-limited hardware such as edge devices and mobile platforms, where power efficiency is a primary concern. 

\begin{table}[h]
\centering
\caption{Comparative Analysis of YOLOv12 with other Object Detection Models}
\label{tab:comp-oo}
\begin{tabular}{|l|c|c|c|c|}
\hline
Model & mAP (\%) & Latency (ms) & FLOPs (G) & Parameters (M) \\
\hline
YOLOv10-N & 38.5 & 1.84 & 6.7 & 2.3 \\
YOLOv11-N & 39.4 & 1.5 & 6.5 & 2.6 \\
\textbf{YOLOv12-N} & \textbf{40.6} & \textbf{1.64} & \textbf{6.5} & \textbf{2.6} \\
RT-DETR-R18 & 46.5 & 4.58 & 60.0 & 20.0 \\
RT-DETRv2-R18 & 47.9 & 4.58 & 60.0 & 20.0 \\
YOLOv11-S & 46.9 & 2.5 & 21.5 & 9.4 \\
\textbf{YOLOv12-S} & \textbf{48.0} & \textbf{2.61} & \textbf{21.4} & \textbf{9.3} \\
\textbf{YOLOv12-M} & \textbf{52.5} & \textbf{4.86} & \textbf{67.5} & \textbf{20.2} \\
YOLOv11-L & 53.3 & 6.2 & 86.9 & 25.3 \\
\textbf{YOLOv12-L} & \textbf{53.7} & \textbf{6.77} & \textbf{88.9} & \textbf{26.4} \\
YOLOv11-X & 54.6 & 11.3 & 194.9 & 56.9 \\
\textbf{YOLOv12-X} & \textbf{55.2} & \textbf{11.79} & \textbf{199.0} & \textbf{59.1} \\
\hline
\end{tabular}
\end{table}

Table \ref{tab:comp-oo} presents a comparative analysis of the YOLOv12 series alongside selected high-performing models from previous YOLO versions and the RT-DETR family. The table showcases key performance metrics including mAP, FLOPs (Giga Floating Point Operations), the number of parameters (Millions), and inference latency (milliseconds). These metrics are directly sourced from the official YOLOv12 paper \cite{tian2025yolov12}, focusing on the models that demonstrate the best performance within their respective categories.

\subsection{Speed Comparison and Hardware Utilization}
The efficiency improvements in YOLOv12 are evident in its superior inference speed and hardware utilization across various platforms. Table \ref{tab:gpu_speed} provides a comparative analysis of inference latency on RTX 3080, RTX A5000, and RTX A6000 GPUs under FP32 and FP16 precision, benchmarking YOLOv12 against YOLOv9 \cite{wang2024yolov9}, YOLOv10 \cite{wang2024yolov10}, and YOLOv11 \cite{yolo11_ultralytics}. For consistency, all experiments were conducted on identical hardware. Furthermore, YOLOv9 and YOLOv10 were evaluated using the Ultralytics codebase \cite{ultralytics_web}.

\begin{table}[h]
\centering
\caption{Performance Comparison of YOLO Models Across GPU Variants \cite{tian2025yolov12}}
\label{tab:gpu_speed}
\renewcommand{\arraystretch}{1.2}
\begin{tabular}{|l|l|l|cc|cc|cc|}
\hline
\textbf{Model} & \textbf{Size} & \textbf{FLOPs (G)} & \multicolumn{2}{c|}{\textbf{RTX 3080}} & \multicolumn{2}{c|}{\textbf{A5000}} & \multicolumn{2}{c|}{\textbf{A6000}} \\
\cline{4-9} 
& & & \textit{FP32} & \textit{FP16} & \textit{FP32} & \textit{FP16} & \textit{FP32} & \textit{FP16} \\
\hline
YOLOv9 [58] & T & 8.2 & 2.4 & 1.5 & 2.4 & 1.6 & 2.3 & 1.7 \\
& S & 26.4 & 3.7 & 1.9 & 3.4 & 2.0 & 3.5 & 1.9 \\
& M & 76.3 & 6.5 & 2.8 & 5.5 & 2.6 & 5.2 & 2.6 \\
& C & 102.1 & 8.0 & 2.9 & 6.4 & 2.7 & 6.0 & 2.7 \\
& E & 189.0 & 17.2 & 6.7 & 14.2 & 6.3 & 13.1 & 5.9 \\
\hline
YOLOv10 [53] & N & 6.7 & 1.6 & 1.0 & 1.6 & 1.0 & 1.6 & 1.0 \\
& S & 21.6 & 2.8 & 1.4 & 2.4 & 1.4 & 2.4 & 1.3 \\
& M & 59.1 & 5.7 & 2.5 & 4.5 & 2.4 & 4.2 & 2.2 \\
& B & 92.0 & 6.8 & 2.9 & 5.5 & 2.6 & 5.2 & 2.8 \\
\hline
YOLOv11 [28] & N & 6.5 & 1.6 & 1.0 & 1.6 & 1.0 & 1.5 & 0.9 \\
& S & 21.5 & 2.8 & 1.3 & 2.4 & 1.4 & 2.4 & 1.3 \\
& M & 68.0 & 5.6 & 2.3 & 4.5 & 2.2 & 4.4 & 2.1 \\
& L & 86.9 & 7.4 & 3.0 & 5.9 & 2.7 & 5.8 & 2.7 \\
& X & 194.9 & 15.2 & 5.3 & 10.7 & 4.7 & 9.1 & 4.0 \\
\hline
YOLOv12 & N & 6.5 & 1.7 & 1.1 & 1.7 & 1.0 & 1.7 & 1.1 \\
& S & 21.4 & 2.9 & 1.5 & 2.5 & 1.5 & 2.5 & 1.4 \\
& M & 67.5 & 5.8 & 1.5 & 4.6 & 2.4 & 4.4 & 2.2 \\
& L & 88.9 & 7.9 & 3.3 & 6.2 & 3.1 & 6.0 & 3.0 \\
& X & 199.0 & 15.6 & 5.6 & 11.0 & 5.2 & 9.5 & 4.4 \\
\hline
\end{tabular}
\end{table}

The results highlight that YOLOv12 significantly outperforms YOLOv9 in inference speed while maintaining comparable efficiency to YOLOv10 and YOLOv11. Notably, on the RTX 3080 GPU, YOLOv12-N achieves an inference time of 1.7 ms (FP32) and 1.1 ms (FP16), marking an improvement over YOLOv9’s 2.4 ms (FP32) and 1.5 ms (FP16). Furthermore, on an NVIDIA T4 GPU, YOLOv12-S achieves an inference latency of 2.61 milliseconds, reinforcing its status as one of the fastest real-time object detection models in its category. This level of efficiency ensures YOLOv12’s viability for latency-sensitive applications.

\begin{figure}[h]
    \centering
    \includegraphics[width=1\linewidth]{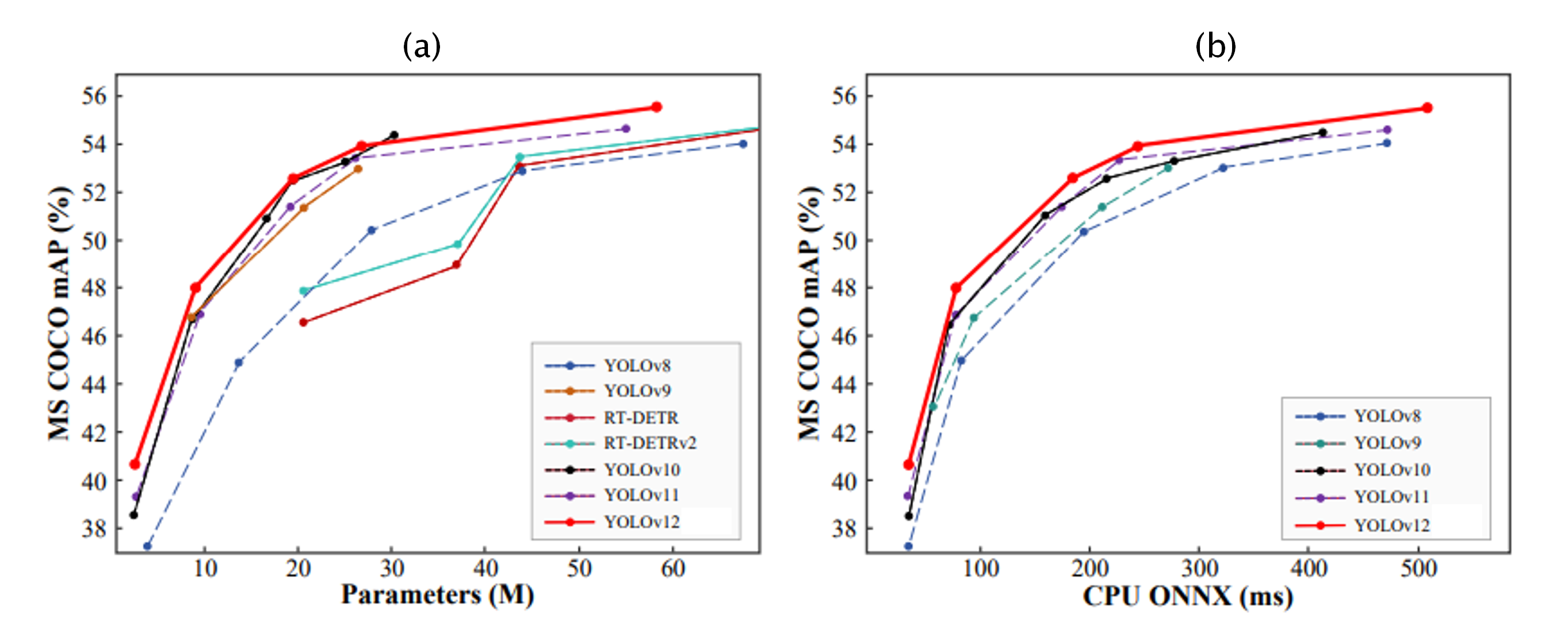}
    \caption{Comparison of YOLOv12 with other SOTA models: (a) accuracy vs. model parameters and (b) accuracy vs. inference latency on CPU \cite{tian2025yolov12}.}
    \label{fig:acc-param-cpu}
\end{figure}

Beyond GPU benchmarks, Figure \ref{fig:acc-param-cpu} provides additional comparative insights into the trade-offs between accuracy, model parameters, and CPU latency. Figure \ref{fig:acc-param-cpu}(a) presents the accuracy-parameter trade-off, where YOLOv12 establishes a dominant boundary, surpassing previous YOLO versions, including YOLOv10, which has a more compact architecture. Figure \ref{fig:acc-param-cpu}(b)  demonstrates accuracy-latency performance on a CPU, where YOLOv12 achieves superior efficiency, surpassing its predecessors when evaluated on an Intel Core i7-10700K @ 3.80GHz.

These improvements are further facilitated by the integration of FlashAttention, which optimizes GPU memory access (SRAM utilization) and reduces memory overhead, enabling higher throughput and lower memory consumption. By addressing bottlenecks in memory access, YOLOv12 allows for larger batch processing and efficient handling of high-resolution video streams, making it particularly well-suited for real-time applications requiring immediate feedback, such as augmented reality, interactive robotics, and autonomous systems.

\section{Key Computer Vision Tasks Supported by YOLO12} \label{sec:CVT}

\subsection{Real-Time Object Detection}  
The YOLO series has consistently prioritized real-time object detection, enhancing the balance between speed and accuracy with each iteration. YOLOv1 introduced the fundamental concept of single-shot detection \cite{redmon2016you}, allowing the model to predict bounding boxes and class probabilities directly from full images in a single evaluation. While groundbreaking in speed, its accuracy suffered from localization errors. YOLOv2 improved upon this by introducing batch normalization, anchor boxes, and multi-scale training, significantly boosting both precision and recall \cite{redmon2017yolo9000}.

Later versions, such as YOLOv3 \cite{redmon2018yolov3} and YOLOv4 \cite{bochkovskiy2020yolov4}, introduced anchor boxes and feature pyramid networks to bolster detection capabilities. Subsequent models, including YOLOv5 and YOLOv6, incorporated optimizations to improve efficiency while maintaining a foundation in convolutional architectures. Notably, YOLOv6 introduced BiC and SimCSPSPPF modules \cite{li2022yolov6}, further refining speed and accuracy. YOLOv7 and YOLOv8 further refined the framework by integrating E-ELAN and C2f blocks for enhanced feature extraction \cite{wang2023yolov7, yolov8_ultralytics}.

YOLOv9 introduced GELAN for architectural optimization and PGI for training improvements \cite{wang2024yolov9}, enabling better gradient flow and increasing robustness against small object detection. YOLOv10 and YOLOv11 shifted towards reducing latency and boosting detection efficiency, with YOLOv11 introducing C3K2 blocks and lightweight depthwise separable convolutions to accelerate detection \cite{khanam2024yolov11}.

Advancing this trajectory, YOLOv12 matches or surpasses its predecessors in real-time performance by integrating attention mechanisms \cite{tian2025yolov12}, previously deemed too slow for such applications. The incorporation of FlashAttention addresses memory bottlenecks, rendering attention processes as swift as traditional convolutional methods while enhancing detection accuracy. Notably, YOLOv12-N achieves a mAP of 40.6\% with an inference latency of 1.64 milliseconds, outperforming both YOLOv10-N and YOLOv11-N in both precision and speed.

\subsection{Object Localization}  
Object localization has been a cornerstone of the YOLO models, with each version refining its bounding box regression capabilities. YOLOv1 initially formulated object detection as a regression problem \cite{redmon2016you}, predicting bounding boxes directly from images without relying on region proposals. However, it lacked anchor-based mechanisms, leading to inconsistent localization accuracy. YOLOv2 introduced anchor boxes and high-resolution classifiers, improving localization precision \cite{redmon2017yolo9000}.

YOLOv3 and YOLOv4 employed anchor-based detection, which, while effective, occasionally resulted in inaccurate bounding boxes due to predefined anchor sizes \cite{redmon2018yolov3, bochkovskiy2020yolov4}. The shift to anchor-free methods and bi-level feature fusion in YOLOv5 and YOLOv6 improved localization accuracy \cite{yolov5_ultralytics, li2022yolov6}. Further optimizations in YOLOv7 and YOLOv8, such as dynamic label assignment \cite{wang2023yolov7} and enhanced loss functions \cite{yolov8_ultralytics}, continued this trend. YOLOv9 enhanced localization by refining feature aggregation strategies and incorporating a more advanced assignment strategy to reduce misalignment \cite{wang2024yolov9}.

YOLOv10 and YOLOv11 introduced improvements in detection heads with C3K2 modules and non-maximum suppression-free (NMS-free) training, refining bounding box predictions \cite{wang2024yolov10, yolo11_ultralytics}. YOLOv12 \cite{tian2025yolov12} enhances object localization by introducing A2, which captures a broader receptive field, leading to more precise localization. The utilization of FlashAttention reduces memory overhead, further improving bounding box regression accuracy, hence surpassing previous versions in localization precision while maintaining rapid inference speeds.

\subsection{Multi-Scale Object Detection}  
The ability to detect objects of varying sizes within the same image has been a focal point of the YOLO series. YOLOv1 and YOLOv2 struggled with small object detection due to limited feature extraction at multiple scales \cite{redmon2016you, redmon2017yolo9000}. YOLOv4 implemented FPN \cite{bochkovskiy2020yolov4} to facilitate multi-scale detection. Enhancements in YOLOv5 and YOLOv6, such as CSPNet \cite{khanam2024yolov5} and SimCSPSPPF \cite{li2022yolov6}, optimized performance across different scales. YOLOv7 and YOLOv8 introduced C2f blocks for improved feature extraction, bolstering multi-scale detection capabilities \cite{wang2023yolov7, yolov8_ultralytics}.

YOLOv9 introduced GELAN, which further improved multi-scale detection by optimizing spatial features across different resolutions \cite{wang2024yolov9}. YOLOv10 and YOLOv11 concentrated on accelerating feature aggregation and employing lightweight detection heads, enhancing performance, particularly for small objects \cite{wang2024yolov10, yolo11_ultralytics}.

YOLOv12 advances multi-scale object detection by incorporating A2 \cite{tian2025yolov12}, which maintains a large receptive field without the need for complex window partitioning, preserving speed. Performance metrics indicate that YOLOv12-N achieves an mAP of 20.2\% for small objects, 45.2\% for medium objects, and 58.4\% for large objects, outperforming previous models across all scales.

\subsection{Optimized Feature Extraction}  
Effective feature extraction is fundamental to object detection, and each YOLO iteration has sought to enhance this process. YOLOv1 relied on fully connected layers, which limited its ability to generalize to unseen object scales \cite{redmon2016you}. YOLOv2 replaced these with deeper convolutional layers and batch normalization, improving efficiency \cite{redmon2017yolo9000}. YOLOv3 and YOLOv4 utilized Darknet-based backbones, which, while powerful, were computationally intensive \cite{redmon2018yolov3, bochkovskiy2020yolov4}.

YOLOv5 and YOLOv6 introduced CSPNet \cite{yolov5_ultralytics} and SimCSPSPPF \cite{li2022yolov6} to optimize feature learning and reduce redundancy. The implementation of E-ELAN and C2f blocks in YOLOv7 and YOLOv8 made feature extraction more efficient \cite{wang2023yolov7, yolov8_ultralytics}. YOLOv9 introduced GELAN, which further optimized the gradient flow and allowed for better utilization of feature maps \cite{wang2024yolov9}.

YOLOv10 and YOLOv11 further improved feature flow with the introduction of C3K2 modules and lightweight convolutions \cite{wang2024yolov10, yolo11_ultralytics}. YOLOv12 introduces the R-ELAN \cite{tian2025yolov12}, enhancing gradient flow and feature integration. The adoption of FlashAttention addresses memory inefficiencies, resulting in faster and more effective feature extraction. These innovations culminate in a superior balance of speed and accuracy, positioning YOLOv12 at the forefront of real-time detection performance.

\subsection{Instance Segmentation}
The evolution of instance segmentation within the YOLO family reflects a shift from simple grid-based detection to high-quality, pixel-level object delineation while maintaining real-time performance.

Early models—YOLOv1, YOLOv2, and YOLOv3—were designed exclusively for bounding box detection and lacked segmentation capabilities \cite{redmon2016you, redmon2017yolo9000, redmon2018yolov3}. A major advancement occurred with YOLOv5, which introduced instance segmentation by incorporating a lightweight, fully convolutional ProtoNet \cite{yolov5_ultralytics}. This enabled the generation of prototype masks that were combined with detection outputs to produce pixel-accurate segmentation masks while retaining high-speed performance.

YOLOv6 focused on architectural improvements such as RepVGG and CSPStackRep blocks, enhancing feature extraction without directly adding a segmentation branch \cite{li2022yolov6}. YOLOv7 introduced a dedicated segmentation variant (YOLOv7-Seg), which preserved real-time efficiency while generating high-quality masks \cite{wang2023yolov7}. YOLOv8 further refined segmentation with an anchor-free segmentation head and an improved backbone, achieving superior accuracy and robust segmentation masks \cite{yolov8_ultralytics}. YOLOv10 introduced adaptive mask resolution, a Feature Alignment Module to reduce mask-box misalignment, and selective transformer elements for capturing long-range dependencies \cite{wang2024yolov10}. These improvements significantly enhanced segmentation quality while maintaining computational efficiency. YOLOv11 optimized segmentation further with the Cross-Stage Partial with Spatial Attention (C2PSA) block, improving focus on relevant regions in cluttered environments \cite{khanam2024yolov11}.

While YOLOv12 does not introduce a dedicated instance segmentation framework, certain architectural enhancements—such as improved attention mechanisms and feature aggregation through R-ELAN—could potentially aid in distinguishing object boundaries more effectively \cite{tian2025yolov12}. FlashAttention, by reducing memory overhead, may also contribute to finer object perception. However, without specific benchmarks or explicit documentation on YOLOv12’s segmentation performance, its advantages in this area remain an area of exploration rather than a confirmed improvement.

\section{Discussion} \label{sec:D}
YOLOv12 represents a substantial advancement in object detection, building upon the strong foundation of YOLOv11 while incorporating cutting-edge architectural enhancements. The model strikes a fine balance between accuracy, speed, and computational efficiency, making it an optimal solution for real-time computer vision applications across diverse domains.

\subsection{Model Efficiency and Deployment}
YOLOv12 introduces a range of model sizes, from nano (12n) to extra-large (12x), allowing for deployment across a variety of hardware platforms. This scalability ensures that YOLOv12 can operate efficiently on both resource-constrained edge devices and high-performance GPUs, maintaining high accuracy while optimizing inference speed. The nano and small variants exhibit significant latency reductions while preserving detection precision, making them ideal for real-time applications such as autonomous navigation \cite{nahavandi2022comprehensive, tang2022perception}, robotics \cite{manakitsa2024review}, and smart surveillance \cite{ghahremannezhad2023object, ramachandran2021review, ahmad2022deep}.

\subsection{Architectural Innovations and Computational Efficiency}
YOLOv12 introduces several key architectural enhancements that improve both feature extraction and processing efficiency. The R-ELAN optimizes feature fusion and gradient propagation, allowing for deeper yet more efficient network structures. Additionally, the introduction of 7×7 separable convolutions reduces the number of parameters while maintaining spatial consistency, leading to improved feature extraction with minimal computational overhead.

One of the standout optimizations in YOLOv12 is the FlashAttention-powered area-based attention mechanism, which enhances detection accuracy while reducing memory overhead. This allows YOLOv12 to localize objects more precisely, especially in cluttered or dynamic environments, without compromising inference speed. These architectural improvements collectively result in higher mAP while maintaining real-time processing efficiency, making the model highly effective for applications requiring low-latency object detection.

\subsection{Performance Gains and Hardware Adaptability}
Benchmark evaluations confirm that YOLOv12 outperforms previous YOLO versions in both accuracy and efficiency. The YOLOv12m variant achieves a comparable or superior mAP to YOLOv11x while using 25\% fewer parameters, showcasing significant computational efficiency improvements. Furthermore, smaller variants, such as YOLOv12s, offer reduced inference latency, making them suitable for edge computing and embedded vision applications \cite{rohith2021comparative}.

From a hardware deployment perspective, YOLOv12 is highly scalable, demonstrating compatibility with both high-performance GPUs and low-power AI accelerators. Its optimized model variants allow for flexible deployment in autonomous vehicles, industrial automation, security surveillance, and other real-time applications \cite{hosain2024synchronizing, hussain2024depth, khanam2025comparative}. The model’s efficient memory utilization and low computational footprint make it a practical choice for environments with strict resource constraints.

\subsection{Broader Implications and Impact}
The innovations introduced in YOLOv12 have wide-reaching implications across multiple industries. Its ability to achieve high-precision object detection with lower computational overhead makes it particularly valuable for autonomous navigation, security, and real-time monitoring systems. Additionally, the model’s small-object detection \cite{iqra2024small} improvements enhance its usability in medical imaging and agricultural monitoring, where detecting fine-grained visual details is critical.

Furthermore, YOLOv12’s efficient processing pipeline ensures seamless deployment across cloud-based, edge, and embedded AI systems, reinforcing its position as a leading real-time detection framework. As the demand for high-speed, high-accuracy vision models continues to rise, YOLOv12 sets a new benchmark in scalable and efficient object detection technology.

\section{Challenges and Future Research Directions} \label{sec:CFR}
Despite YOLOv12’s architectural advancements and efficiency, several challenges remain that warrant further research. Addressing these limitations will be crucial for optimizing deployment in real-world applications and expanding YOLOv12’s capabilities beyond standard object detection.  

\subsection{Hardware Constraints and Deployment on Edge Devices}
While YOLOv12 integrates attention mechanisms and FlashAttention to improve accuracy, these enhancements come with increased computational demands. Although the model achieves real-time performance on high-end GPUs, deploying it on low-power edge devices such as mobile processors, embedded systems, and IoT devices remains a challenge \cite{ajani2021overview}.  

One key limitation is memory bottlenecks. Attention-based architectures require higher VRAM usage due to extensive feature maps and matrix multiplications. This makes it difficult to run YOLOv12 efficiently on resource-constrained devices such as NVIDIA Jetson Nano, Raspberry Pi, and ARM-based microcontrollers \cite{iqbal2024review}. Optimizing memory footprint through model compression techniques like low-rank decomposition \cite{saha2025compressing} and weight pruning \cite{naveen2022memory} could help alleviate this issue.  

Another challenge is inference latency. While YOLOv12 reduces attention overhead compared to full Vision Transformers \cite{fang2024eva, he2022masked}, it still lags behind pure CNN-based YOLO versions on edge hardware. Strategies such as structured pruning, knowledge distillation, and quantization (e.g., int8) could improve real-time performance on embedded AI accelerators \cite{alhussain2024efficient}.  

Additionally, future research could explore hardware-specific optimizations to enhance YOLOv12’s efficiency across diverse platforms. Techniques such as tensor-level optimizations \cite{huang2024fasor}, efficient convolutional kernels \cite{guo2022efficient}, and FPGA/DSP implementations could make the model more adaptable for low-power devices \cite{garcia2014survey}.  

\subsection{Training Complexity and Dataset Dependency}
The improvements in YOLOv12’s accuracy come at the cost of increased training complexity and higher dataset dependency. Unlike earlier YOLO models that were optimized for lightweight training, YOLOv12 introduces attention mechanisms and deeper feature aggregation, which result in higher computational requirements.  

One major challenge is training cost. Attention-based modules require significantly more FLOPs and memory bandwidth, making training expensive, especially for researchers with limited GPU resources. Techniques like low-rank factorization of attention weights, gradient checkpointing, and efficient loss functions could help reduce computational overhead \cite{mei2025optimizing}.  

Another issue is data efficiency. YOLOv12's superior accuracy is largely due to training on large-scale datasets like MS COCO and OpenImages. However, in many real-world applications such as medical imaging \cite{sarvamangala2022convolutional} and industrial defect detection \cite{khanam2024comprehensive}, datasets are often small or imbalanced. Exploring self-supervised learning, semi-supervised training, and domain adaptation techniques \cite{rani2023self, yang2022survey, shirdel2022survey} could improve YOLOv12’s performance in low-data environments.  

Furthermore, hyperparameter sensitivity remains a challenge. YOLOv12 requires extensive tuning of parameters like learning rates, attention heads, and anchor box sizes, which can be computationally expensive. Future research could investigate automated hyperparameter tuning using techniques like NAS \cite{elsken2019neural} to improve usability and efficiency.  

\subsection{Expanding Beyond Object Detection}
While YOLOv12 is optimized for 2D object detection, many emerging applications require more advanced scene understanding beyond simple bounding boxes. Expanding YOLOv12 into 3D object detection, instance segmentation, and panoptic segmentation could open new research opportunities.  

For 3D object detection, applications like autonomous driving \cite{mao20233d} and robotics \cite{wong1998robotic} require models that can predict depth-aware 3D bounding boxes. Current transformer-based models like DETR3D and BEVFormer leverage multi-view inputs and LiDAR fusion \cite{zhong2023transformer}. Extending YOLOv12 to process stereo images or LiDAR data could make it suitable for 3D perception tasks.  

For instance segmentation, YOLOv12 lacks a dedicated segmentation head. Existing solutions like YOLACT and SOLOv2 enable real-time instance segmentation by integrating lightweight mask branches \cite{yang2023review}. Future iterations of YOLO could incorporate a parallel segmentation branch to improve pixel-wise object delineation.  

Moreover, panoptic segmentation \cite{elharrouss2021panoptic}, which combines instance and semantic segmentation, has become a growing area in computer vision. While current YOLO models do not support this task, integrating transformer-based segmentation heads while maintaining YOLO’s efficiency could enable a unified object detection and segmentation framework.  

\section{Conclusion} \label{sec:C}
In this review, we have presented an in-depth analysis of YOLOv12, the latest evolution in the YOLO family of real-time object detectors. By integrating innovative techniques such as the A2 module, R-ELAN, and FlashAttention, YOLOv12 effectively balances the trade-off between accuracy and inference speed. These enhancements not only address the limitations inherent in earlier YOLO versions and traditional convolutional approaches but also push the boundaries of what is achievable in real-time object detection.

We have traced the technical evolution of YOLO architectures and detailed the structural refinements in YOLOv12, including its optimized backbone and detection head. Comprehensive benchmark evaluations demonstrate that YOLOv12 achieves superior performance across multiple metrics, including latency, accuracy, and computational efficiency, making it well-suited for both high-performance GPUs and resource-constrained devices.

While YOLOv12 marks a significant advancement, our review also identifies several challenges that remain, such as hardware constraints for edge deployment and training complexity. Overall, YOLOv12 represents a substantial step forward in real-time object detection, combining the strengths of convolutional and attention-based approaches. Its scalable design and enhanced efficiency not only cater to a wide range of applications but also pave the way for further innovations in computer vision.

%%%%%%%%%%%%%%%%%%%%%%%%%%%%%%%%%%%%%%%%%%
\vspace{6pt} 

%%%%%%%%%%%%%%%%%%%%%%%%%%%%%%%%%%%%%%%%%%

\bibliographystyle{unsrt}  % Changes bibliography style to unsorted
\bibliography{ref}  % This points to the filename of your BibTeX file without the .bib extension

\end{document}